\documentclass{article}

\usepackage{PRIMEarxiv}

\usepackage[utf8]{inputenc} 
\usepackage[T1]{fontenc}    
\usepackage{hyperref}       
\usepackage{url}            
\usepackage{booktabs}       
\usepackage{amsmath}        
\usepackage{amsfonts}       
\usepackage{nicefrac}       
\usepackage{microtype}      
\usepackage{lipsum}
\usepackage{fancyhdr}       
\usepackage{graphicx}       
\graphicspath{{media/}}     

\title{TopoMamba: A Load-Support Relation-Guided Multi-Directional State-Space Model for Topology Optimization

}

\author{
  Bin Lou, Yuxuan Cheng, Huaizhi Zong, Junhui Zhang, Bing Xu \\
  State Key Laboratory of Fluid Power and Mechatronic Systems\\
  Zhejiang University\\
  Hangzhou\\
  \texttt{\{binlou, cyx5554, hzzong, benzjh, bxu\}@zju.edu.cn} \\
 \\
}

\begin{document}
\maketitle

\begin{abstract}

Deep learning has emerged as an efficient alternative for predicting high-performance material distributions in topology optimization. Existing methods struggle to accurately capture load-transfer information, limiting out-of-distribution generalization, while their model architectures often incur high computational costs. To address these challenges, this paper proposes TopoMamba, a topology prediction framework  incorporating a load-support relation-guided multi-directional state-space model. Coupling physical fields with load-support relations enables more effective modeling of mechanical dependencies. A load-support relation-guided spatially adaptive fusion mechanism dynamically adjusts multi-directional scan features according to spatial conditions. Mamba is coupled with the solid isotropic material with penalty method to enhance structural mechanical performance while maintaining computational efficiency. Results on two-dimensional topology optimization benchmarks demonstrate that TopoMamba achieves superior topology prediction accuracy, out-of-distribution generalization, and computational efficiency over state-of-the-art models. The proposed load-support physics-guided framework enables efficient optimization of more complex structural systems.

\end{abstract}


\section{Introduction}
As an advanced design method, topology optimization (TO) aims to obtain high-performance structures by optimizing the distribution of materials under the constraints of load, boundary conditions and material volume\cite{bendsoe1989material,bendsoe1988homogenization}. Gradient-based TO repeats the finite element analysis and material distribution update during the iterative process, leading to high computational cost. With the increase of grid resolution, the computational cost of TO rises sharply, which limits its widespread application \cite{TRAFF2023116043}.

Deep learning has been widely employed to reduce the computational cost of TO by learning an end-to-end mapping from design conditions to optimal material distributions\cite{woldseth2022ann,shin2023review,sosnovik2019neural}. The existing end-to-end methods include generative methods and deterministic methods. Generative methods offer strong distribution modeling capabilities but often incur high computational costs due to multi-step sampling or iterative refinement\cite{ho2020denoisingdiffusionprobabilisticmodels,song2022denoisingdiffusionimplicitmodels}. Deterministic methods enable efficient prediction through a single forward pass, but their generalization to unseen conditions is often limited by the training distribution. Therefore, improving generalization while retaining computational efficiency remains an important challenge.

Achieving this balance requires addressing several limitations in mechanical condition representation, long-range dependency modeling, and computational efficiency. Although the existing methods enhance the characterization of  conditions by introducing physical fields\cite{nie2021topologygan,maze2023topodiff}, it is still difficult to fully reflect the mechanical information contained in loads, boundary conditions and their relations. Moreover, the material distribution in TO has a significant spatial dependence. The local material distribution is not only affected by the neighborhood response, but also related to the long-distance load, support and overall force-transfer process. Meanwhile, global self-attention captures long-range dependencies but incurs quadratic computational and memory costs with respect to the number of tokens \cite{vaswani2017attention}.

To solve these problems, this paper proposes TopoMamba, a multi-directional state-space topology prediction framework guided by load-support relationship. We combine the physical field and the explicitly constructed load-support relationship to accurately characterize the information of conditions. We further adaptively reweight directional state-space features based on spatial position and load-support configuration, enabling long-range information propagation to better follow force-transfer patterns. Finally, the efficient deterministic prediction and the solid isotropic material with penalty (SIMP) method are combined to improve the mechanical properties of the generated structure while maintaining a low computational cost.

Our main contributions are summarized as follows:(1) an explicit load-support relation representation to enhance the modeling of global mechanical dependencies in topology prediction.
(2)  a load-support relation-guided spatially adaptive multi-directional state-space model, enabling long-range feature modeling to adapt to different mechanical conditions.
(3)  an efficient prediction-optimization collaborative topology design framework that further improves structural mechanical performance while achieving a favorable balance between structural quality and computational efficiency.

\section{Related Work}
\subsection{Classical Topology Optimization}
\begin{figure}[!htb]
    \centering
    \includegraphics[width=0.8\linewidth]{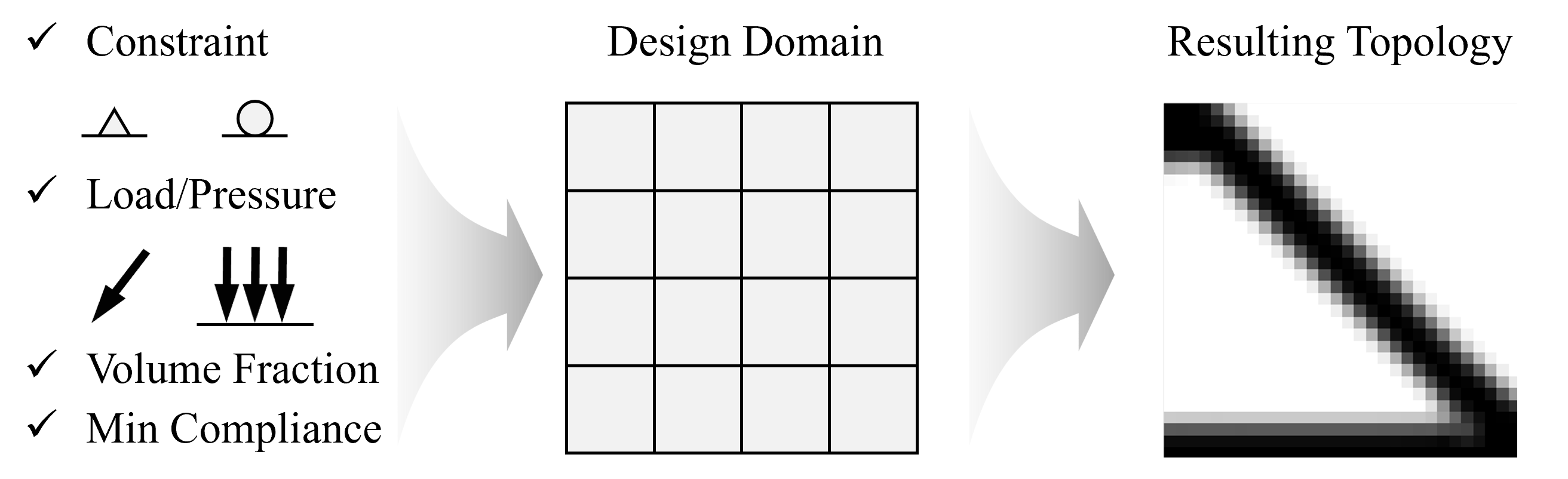}
    \caption{TO. TO introduces the design domain, load and support conditions, and volume fraction as input conditions. The optimization process iteratively updates the material distribution to minimize compliance while satisfying the volume fraction constraint.}
    \label{fig:TO}
\end{figure}

As shown in Figure \ref{fig:TO}, classical topology optimization aims to obtain high-performance structures by optimizing material distributions under prescribed design, loading, boundary, and material constraints, with objectives such as minimizing structural compliance, thermal compliance, or pressure loss.\cite{bendsoe1989material, sigmund2013review}. Among them, SIMP is one of the most widely used density methods in structural TO \cite{sigmund2001code, andreassen2011code}. SIMP relaxes the existence of discrete materials into continuous element density variables, and establishes the relationship between element density and material stiffness through the material interpolation model with penalty. The structural response and the sensitivity of the objective function are obtained according to the finite element analysis, and the gradient-based optimization algorithms such as the optimality criteria or the method of moving asymptotes \cite{svanberg1987mma} are used to iteratively update the material distribution. Although the classical TO can obtain the material distribution with excellent mechanical properties, the solution process usually needs to re perform the finite element analysis and sensitivity calculation after each design variable update. Therefore, with the increase of design freedom and problem complexity, a large number of repeated numerical solutions will bring higher computational costs.

\subsection{Deep Learning in TO}

The application of deep learning in TO is mainly carried out along two paths: one is to use neural network to approximate finite element analysis, so as to reduce the repeated calculation cost \cite{woldseth2022ann}; The other class directly learns the mapping between the design conditions and the optimal material distribution to achieve fast end-to-end topology prediction. In recent years, end-to-end methods have gradually become an important direction of learning TO, and have developed different prediction paradigms based on Gan, diffusion model and neural implicit representation.\cite{nie2021topologygan, maze2023topodiff, bastek2025pidm, nobari2025nito, yang2026hpgdiff, ZHENG2021107263}.

Learning-based TO is gradually changing from only using design conditions such as load, boundary conditions and volume fraction to further using physical fields. Topologygan \cite{nie2021topologygan} takes the physical field in the initial design domain as conditional information to enhance topology prediction; Topodiff \cite{maze2023topodiff} further combines conditional diffusion and physical performance to guide the generation of topology that meets the design requirements; Recent topotransformer \cite{rashed2026topotransformer} further shows that the physical fields related to sensitivity plays an important role in generalization without working conditions. However, physical fields can not fully capture the mechanical interactions among loads and support conditions.

\subsection{State-space Models}

Transformer relies on self attention mechanism to achieve global feature interaction and can effectively model long-range dependencies \cite{vaswani2017attention, dosovitskiy2021vit}. However, its computational and storage overhead usually increases with the square of the sequence length, which will bring a large computational burden in high-resolution topology prediction. The state-space model provides another efficient paradigm for long-range dependency modeling \cite{gu2022s4}. Mamba introduces a selective mechanism based on the structured state-space model, which enables the model to dynamically control the retention and propagation of information according to the input content, and reduces the complexity of sequence modeling to a linear scale with respect to sequence length \cite{yang2024plainmamba,li2024mamband}. Subsequently, vision Mamba and others further extended the mechanism to two-dimensional visual tasks, enabling the state-space model to realize large-scale spatial feature interaction at a low computational cost \cite{zhu2024visionmamba}. Vmamba further propagates and aggregates features along multiple spatial directions through two-dimensional selective scanning, enabling different locations to obtain more sufficient long-range context information \cite{liu2024vmamba}.In this study, mamba is applied to the field of TO, and modified based on the TO task, without attention mechanism.

\section{Method}
\subsection{Preliminaries}
\subsubsection{Topology Optimization}
The TO problem can be written as follows:
\begin{equation}\label{eq:to}
\begin{aligned}
\min_{\boldsymbol{\rho}}:\quad
& c(\boldsymbol{\rho})
= \mathbf{U}^{T}\mathbf{K}(\boldsymbol{\rho})\mathbf{U}
\\
\text{subject to:}\quad
& \frac{V(\boldsymbol{\rho})}{V_0}\leq v,
\\
& \mathbf{K}(\boldsymbol{\rho})\mathbf{U}=\mathbf{F},
\\
& \rho_{\min}\leq \rho_e\leq 1,
\qquad e=1,\ldots,N.
\end{aligned}
\end{equation}
where $c$ denotes the structural compliance, $\mathbf{U}$ is the global displacement vector, $\mathbf{K}$ is the global stiffness matrix, $\mathbf{F}$ is the global force vector, and $\mathbf{\rho}$ is the element density vector. The volume constraint requires the ratio of the material volume $V(\boldsymbol{\rho})$ to the design domain volume $V_0$ to be no greater than the target volume fraction $v$. The density variable $\rho_e$ for each element $e$ is bounded between $\rho_{\min}$ and 1, and $N$ is the total number of elements in the design domain.

Based on SIMP method, the Young's modulus of element \(e\) is interpolated as
\begin{equation}\label{eq:simp}
E_e(\rho_e)
=
E_{\min}
+
\rho_e^{p}\left(E_0-E_{\min}\right),
\end{equation}
where \(E_0\) denotes the Young's modulus of the solid material, $E_{\min}$ is a small stiffness introduced to avoid singularity of the global stiffness matrix, and $p$ is the penalization factor to discourage intermediate densities.

\subsubsection{State-Space Models and Mamba}
A continuous linear state-space model maps an input $x(t)$ to an output $y(t)$ through a hidden state $\mathbf h(t)$:
\begin{equation}\label{eq:ssm}
\dot{\mathbf h}(t)=\mathbf A\mathbf h(t)+\mathbf Bx(t),\qquad
y(t)=\mathbf C\mathbf h(t)+ \mathbf D x(t).
\end{equation}
where \(\mathbf A\), \(\mathbf B\), \(\mathbf C\), and \(\mathbf D\) are the state-space parameters. Mamba introduces a selective mechanism with input-dependent parameters, enabling adaptive information propagation along a sequence. For two-dimensional visual data, image features can be organized into sequences along spatial scanning paths, allowing Mamba to capture long-range spatial dependencies.

\subsection{Architecture}

\begin{figure}[t]
    \centering
    \includegraphics[width=\linewidth]{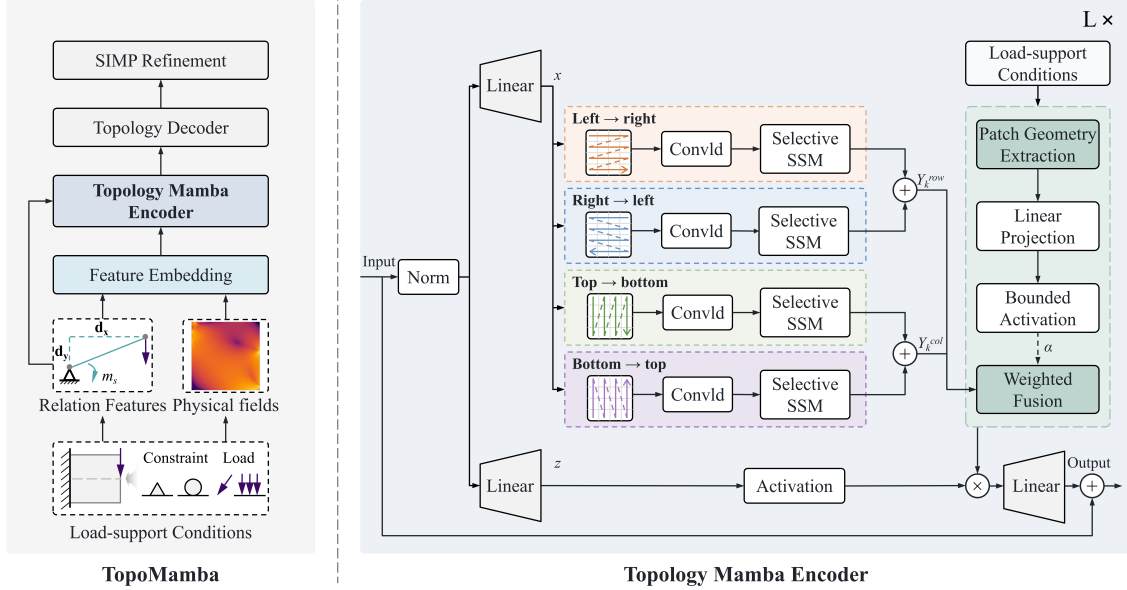}
    \caption{Overview of the proposed TopoMamba framework. The framework illustrates the complete workflow from prescribed loading and boundary conditions to SIMP refinement. The encoder models features along four spatial directions using selective state-space modules and adaptively fuses the directional features with spatial weights conditioned on the load-support relations.}
    \label{fig:architecture}
\end{figure}

The overall network architecture of TopoMamba is illustrated in Figure \ref{fig:architecture}. It adopts an encoder-decoder architecture that embeds the input conditions into spatial features, processes them with stacked state-space blocks, and decodes them into a continuous material-density field, which is subsequently refined by SIMP.

Given the design conditions $\mathbf{c}$, we first construct two complementary condition representations: a physics-conditioned representation $\mathbf{X}$ and an explicit load--support relation representation $\mathbf{R}$. These representations are subsequently encoded and fused into an element-wise feature map:
\begin{equation}
\mathbf{S}=\mathcal{E}_{\theta}(\mathbf{X},\mathbf{R}),
\label{eq:input_encoding}
\end{equation}
where $\mathbf{S}\in\mathbb{R}^{H\times W\times C}$ denotes the resulting spatial feature representation.

To construct the input sequence for the state-space backbone, 
$\mathbf{s}$ is partitioned into non-overlapping $P\times P$ patches.
Each patch is flattened and linearly projected into a $D$-dimensional token.
The resulting token sequence is
\begin{equation}
\mathbf{T}_0 =
\left[
\mathbf{s}_p^1 \mathbf{W}_p;
\mathbf{s}_p^2 \mathbf{W}_p;
\ldots;
\mathbf{s}_p^J \mathbf{W}_p
\right]
+
\mathbf{E}_{\mathrm{pos}},
\label{eq:patch_embedding}
\end{equation}
where $\mathbf s_p^j\in\mathbb R^{P^2C}$ denotes the $j$-th flattened patch,
$\mathbf W_p\in\mathbb R^{(P^2C)\times D}$ is the learnable projection matrix,
$J=HW/P^2$ is the number of patches, and
$\mathbf E_{\mathrm{pos}}\in\mathbb R^{J\times D}$ is the positional embedding.
The token sequence is then processed by the state-space backbone:
\begin{equation}\label{eq:architecture}
\begin{aligned}
\mathbf T_l
&=
\operatorname{Mamba}
\left(
\mathbf T_{l-1};
\boldsymbol{\alpha}(\mathbf R)
\right)+\mathbf T_{l-1},\\
\widehat{\boldsymbol{\rho}}
&=
\operatorname{sigmoid}
\left[
-\operatorname{Unpatchify}
\left(
\operatorname{Linear}
\left(
\operatorname{LN}(\mathbf T_l)
\right)
\right)
\right],
\end{aligned}
\end{equation}
where $\boldsymbol{\alpha}(\mathbf R)$ contains the fusion weights conditioned on the design problem, $L$ is the number of mamba blocks, and $\mathbf T_l$ is the output of the $l$-th layer of the mamba block.
The network is trained with a composite loss function that combines a binary cross entropy loss $\mathcal L_{\mathrm{BCE}}$ between the predicted and reference topologies, volume fraction loss $\mathcal L_{\mathrm{vol}}$, and the floating material loss $ \mathcal L_{\mathrm{floating}}$:
\begin{equation}
\mathcal L
=\mathcal L_{\mathrm{BCE}}
+\lambda_v\mathcal L_{\mathrm{vol}}
+\lambda_f\mathcal L_{\mathrm{floating}}.
\end{equation}
Here, \(\lambda_v\) and \(\lambda_f\) denote the weighting coefficients for the volume fraction and floating material losses, respectively.

\subsection{Load-Support Relation Embedding}
\label{sec:input_encoding}

Following prior learning-based methods \cite{nie2021topologygan,maze2023topodiff}, we adopt strain energy density
 and von Mises stress as  physical-field inputs obtained from
finite-element analysis. While these fields characterize the structural response
under the prescribed conditions, they do not explicitly represent the spatial
relation between the load and supports. We therefore construct
the explicit load-support relation representation.

With coordinates normalized to the unit domain, let
$\boldsymbol{\ell}$, $\mathbf{f}$, and $\mathbf{s}$ denote the load position,
applied force, and support position, respectively. We define
\begin{equation}
\begin{aligned}
\mathbf{d}_s &= \boldsymbol{\ell}-\mathbf{s},
& \widehat{\mathbf{f}} &= \frac{\mathbf{f}}{\|\mathbf{f}\|_2}, \\
m_s &= d_{s,x}\widehat{f}_y-d_{s,y}\widehat{f}_x,
& r_s^2 &= \|\mathbf{d}_s\|_2^2 .
\end{aligned}
\label{eq:relations}
\end{equation}
For each support node $\mathbf{s}$, let $b_x(\mathbf{s})$ and
$b_y(\mathbf{s})$ indicate whether the node is constrained along the
$x$- and $y$-directions, respectively. The load--support relation feature is
defined as
\begin{equation}
\mathbf{R}(\mathbf{s}) =
\left[
b_x(\mathbf{s})\,\mathbf{r}(\mathbf{s});
b_y(\mathbf{s})\,\mathbf{r}(\mathbf{s})
\right],
\label{eq:relationmap}
\end{equation}
where
\begin{equation}
\mathbf{r}(\mathbf{s}) =
\left[
1,\,
d_{s,x},\,
d_{s,y},\,
\widehat f_x,\,
\widehat f_y,\,
m_s,\,
r_s^2
\right].
\end{equation}

\subsection{Load-Support Relation-Guided Scan Fusion}
\label{sec:scan_fusion}

The material distribution in TO is strongly influenced by
the relative positions of loads and supports. Since their effects can vary
across spatial directions under different design conditions, uniformly
combining directional features may be suboptimal. We therefore use the
load--support relations introduced in Sec.~\ref{sec:input_encoding} to
adaptively fuse multi-directional state-space features.

Let $\mathbf{T}_{k-1}$ denote the input feature grid of the $k$-th TopoMamba
block. We apply selective state-space scans along four spatial directions,
$\{\rightarrow,\leftarrow,\downarrow,\uparrow\}$. Opposite directions are
aggregated into row-wise and column-wise representations:
\begin{equation}
\mathbf{Y}_{k}^{\mathrm{row}}
=
\mathcal{S}_{k}^{\rightarrow}(\mathbf{T}_{k-1})
+
\mathcal{S}_{k}^{\leftarrow}(\mathbf{T}_{k-1}),
\qquad
\mathbf{Y}_{k}^{\mathrm{col}}
=
\mathcal{S}_{k}^{\downarrow}(\mathbf{T}_{k-1})
+
\mathcal{S}_{k}^{\uparrow}(\mathbf{T}_{k-1}),
\end{equation}
where $\mathcal{S}_{k}^{d}$ denotes the selective state-space scan along
direction $d$, with all outputs aligned to their original patch locations.

For each patch position $\mathbf q$, we extract local relation features
$\mathbf g(\mathbf q;\mathbf R)$ from the load--support representation
$\mathbf R$ introduced in Sec.~\ref{sec:input_encoding}. A linear projection followed by a bounded activation produces a spatially varying fusion weight \(\alpha_k(q)\in[0,1]\).
The final representations are then combined as
\begin{equation}
\mathbf{Y}_k(\mathbf q)
=
\alpha_k(\mathbf q;\mathbf R)
\mathbf{Y}_{k}^{\mathrm{row}}(\mathbf q)
+
\left[1-\alpha_k(\mathbf q;\mathbf R)\right]
\mathbf{Y}_{k}^{\mathrm{col}}(\mathbf q).
\end{equation}

Thus, the state-space scan paths remain fixed, while their relative
contributions vary across spatial locations according to the load--support
configuration.

\subsection{SIMP Refinement}\label{sec:refinement}

TopoMamba predicts a continuous material-density field, while the final topology is binary. Since structural compliance is not directly optimized by the loss, the prediction may still contain suboptimal regions; moreover, direct thresholding introduces a mismatch between the continuous representation used during training and the discrete topology required at inference. To address both issues, we apply a small number of SIMP iterations before binarization, using compliance sensitivities and the prescribed volume constraint to refine the predicted density field in the continuous design space. This refinement provides local  corrections while preserving the efficiency of the learned prediction.
\begin{equation}
\boldsymbol{\rho}^{(0)}=\widehat{\boldsymbol{\rho}}, \qquad
\boldsymbol{\rho}^{(j+1)}
=
\mathcal{U}_{\mathbf c}\!\left(\boldsymbol{\rho}^{(j)}\right),
\quad j=0,\ldots,K-1,
\end{equation}
each update uses the current structural response and compliance sensitivities to correct mechanically important regions under the prescribed volume constraint. Binarization is applied only after refinement, so the network prediction and the physics-based updates remain in the same continuous design space. 
\section{Experiments}
In this section, we evaluate the performance of TopoMamba on two-dimensional structural TO benchmarks. We compare TopoMamba with state-of-the-art deterministic and generative models in terms of compliance error, volume fraction error, and end-to-end runtime. We also conduct ablation studies to assess the contributions of key  components.
\subsection{Experimental Setup}
\subsubsection{Dataset and Baselines}
We use the two-dimensional structural TO benchmarks from TopoDiff and NITO. The dataset includes 30,000 training samples and 1,800 in-distribution (ID) conditions samples for testing our models. Additionally, we include a dataset of 1,000 out-of-distribution (OOD) samples for generalization testing. We compare TopoMamba with the baseline of deterministic models and generative models. 
\subsubsection{Evaluation Metrics}
In line with NITO and TopoDiff, we evaluate the models using three metrics, as follows:

\textbf{Compliance Error.}
    Structural performance is evaluated using the compliance error (CE), which compares the compliance of the predicted topology with that of a reference design obtained by SIMP under the same loading and boundary conditions. For the $i$-th sample, CE is defined as
    \begin{equation}
        \mathrm{CE}_i
        = 
        \frac{C(\hat{y}_i)-C(y_i)}{C(y_i)}
        ,
    \end{equation}
    where $C(\hat{y}_i)$ denotes the compliance of the predicted topology and $C_i^{\mathrm{ref}}$ denotes the compliance of the corresponding SIMP-optimized reference design. We report both the mean and median CE to evaluate the compliance performance of each model.

\textbf{Volume Fraction Error.}
    We further evaluate constraint satisfaction using the volume fraction error(VFE), which quantifies the relative deviation between the actual volume fraction of the generated topology and the prescribed target volume fraction:
    \begin{equation}
        \mathrm{VFE}_i
        = \frac{
        \left| VF(\hat{y}_i) - VF(y_i) \right|
        }{
        VF(y_i)
        },
    \end{equation}
where, $VF(\hat{y}_i)$ denotes the actual volume fraction of the predicted topology, while $VF(y_i)$ denotes the target volume fraction.

\textbf{End-to-End Runtime.}
We measure the end-to-end runtime required by each method to generate the final topology from a given problem specification. For comparison, we also report the runtime of the standard SIMP optimizer under the same problem setting. All experiments are conducted on an NVIDIA RTX 5880 Ada 48 GB GPU and an AMD EPYC 9654 CPU.

We exclude samples with compliance errors exceeding $1000\%$, consistent with a common practice in prior SOTA models  (Mazé and Ahmed, 2023; Giannone et al., 2023; Nie et al., 2021b) for comparsion purposes. Such cases typically correspond to failed predictions in which the model does not produce a meaningful material distribution or satisfy the prescribed constraints.

\subsection{Performance}

\begin{table}[!htb]
\caption{Performance comparison on ID tests. FS: the number of SIMP refinement steps. w/ G: using a classifier and regression guidance. HPG-Diff* and TopoViT* results from \cite{yang2026hpgdiff} and \cite{lutheran2026physicsinformedtransformerrealtimehighfidelity}}.
\label{tab:id}\centering
\begin{tabular}{lcccc}\hline
Method & FS & Mean CE (\%) & Median CE (\%) & VFE (\%)\\\hline
TopologyGAN & -- & 48.51 & 2.06 & 11.87\\
TopoDiff & -- & 3.23 & 0.45 & 1.14\\
TopoDiff w/ G & -- & 2.59 & 0.49 & 1.18\\
HPG-Diff* & -- & 0.87 & 0.16 & 1.38\\\hline
NITO & 5 & 0.81 & 0.098 & 0.5\\
NITO & 10 & 0.48 & 0.067 & 0.41\\
TopoViT* & -- & 3.67 & 0.91 & 4.03\\
TopoMamba & -- & 2.91 & 0.27 & 0.96\\
TopoMamba & 5 & 0.43 & 0.13 & 0.50\\
TopoMamba & 10 & $\mathbf{0.17}$ & 0.074 & $\mathbf{0.29}$ \\\hline
\end{tabular}
\end{table}
\begin{figure}[t]
    \centering
    \includegraphics[width=\linewidth]{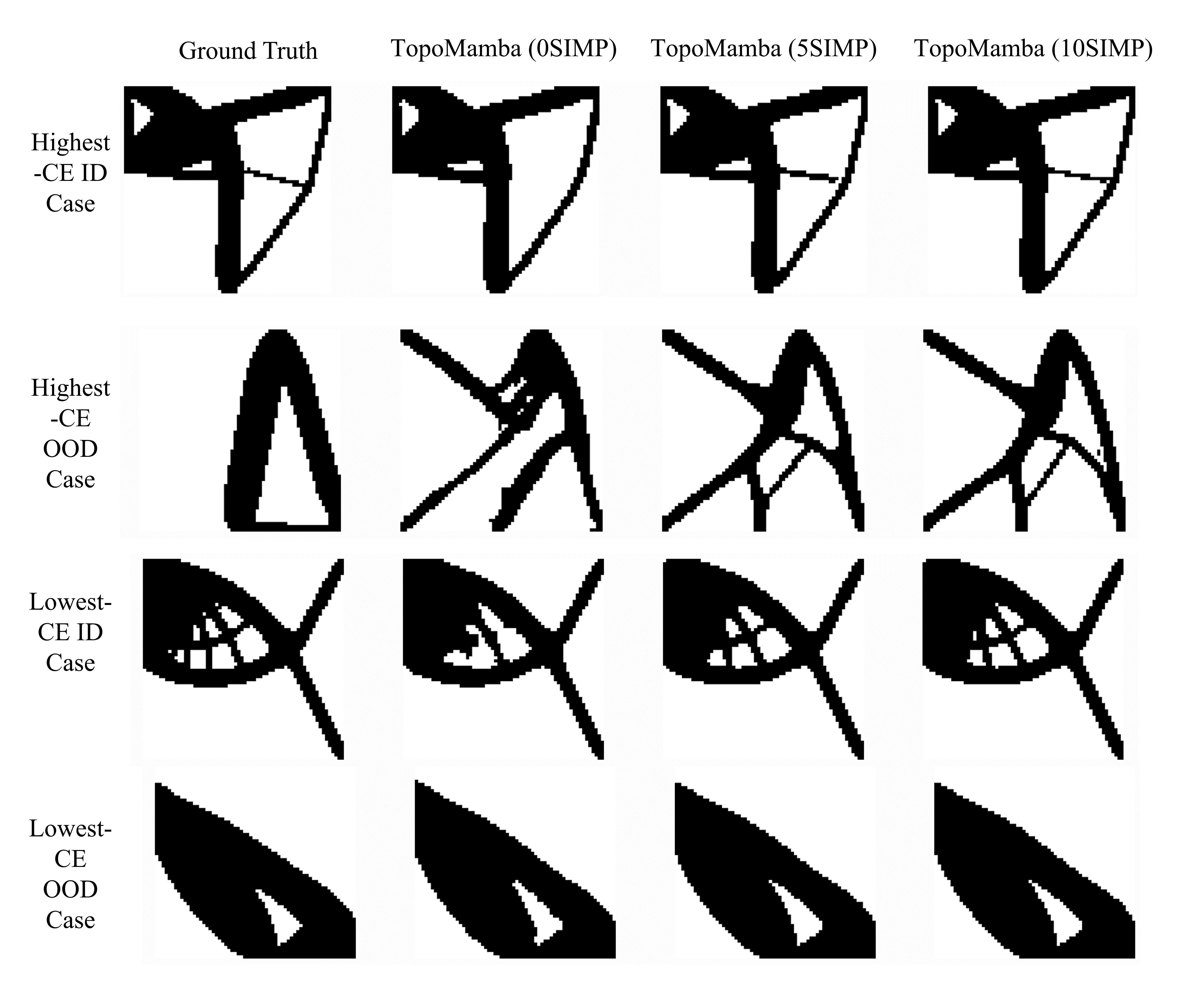}
    \caption{Qualitative comparison of TopoMamba before and after SIMP refinement on representative ID and OOD cases.}
    \label{fig:qualitative}
\end{figure}
\begin{table}[!htb]
\caption{Performance comparison on OOD tests. }
\label{tab:ood}\centering
\begin{tabular}{lcccc}\hline
Method & FS & Parameters (M) &Mean CE (\%)&  Median CE (\%)\\\hline
TopoDiff & -- & 121 & 8.57 & 1.14\\
TopoDiff w/ G & -- & 239 & 7.79 & 1.26\\
TopoTransformer & -- & 34 & 5.73 & 0.53\\
NITO & 5 & 22 & 9.33 & 2.37\\
TopoMamba & 5 & $\mathbf{7}$ & 1.79 & 0.28\\
TopoMamba & 10 & $\mathbf{7}$ & $\mathbf{1.07}$ & $\mathbf{0.14}$\\
HPG-Diff & -- & 177 & 5.29 & 0.61\\\hline
\end{tabular}
\end{table}
Table \ref{tab:id} presents the performance of different methods on ID tests . TopoMamba achieves low errors in terms of mean compliance error, median compliance error, and volume fraction error, demonstrating favorable mechanical performance and constraint satisfaction. Its best results reach 0.17\% Mean CE, 0.074\% Median CE, and 0.29\% VFE, respectively. Notably, even without SIMP refinement, TopoMamba already achieves performance comparable to several existing state-of-the-art methods. Further refinement leads to additional improvements across the evaluation metrics, indicating both the high quality of the initial designs generated by TopoMamba and their potential for further optimization.

Table \ref{tab:ood} shows the performance of different methods on OOD tests. TopoMamba achieves state-of-the-art performance while using only 7M parameters, demonstrating strong generalization capability and high parameter efficiency. Its best results reach 1.07\% Mean CE and 0.14\% Median CE, outperforming the compared methods on both metrics. These results highlight the effectiveness of TopoMamba in maintaining accurate and robust performance under distribution shifts, while requiring substantially fewer model parameters.

Figure \ref{fig:qualitative} presents qualitative comparisons of TopoMamba before and after SIMP refinement on representative ID and OOD cases. The results demonstrate that TopoMamba can generate high-quality initial topologies that closely resemble the reference designs, even under challenging OOD conditions. The subsequent SIMP refinement further enhances the structural quality, yielding final designs with improved compliance and adherence to volume constraints. These visualizations underscore the effectiveness of TopoMamba in producing mechanically sound and visually coherent topologies across diverse scenarios.
\paragraph{End-to-End Runtime.}
Table \ref{tab:runtime} compares the end-to-end runtime for obtaining the final topology from a given problem specification, including physical field preparation, model inference, and numerical refinement. TopoMamba achieves an average end-to-end runtime of 0.48 s, substantially faster than TopoTransformer and HPG-Diff, while remaining close to the fastest learning-based method, NITO. This demonstrates that TopoMamba provides high computational efficiency in addition to its strong predictive performance. 
\begin{table}[!htb]
\caption{Averaged End-to-end times of tests.}
\label{tab:runtime}\centering\small
\begin{tabular}{lr}\hline Method & Time (s)\\\hline
Standard SIMP & 10.59\\
NITO & 0.40\\
TopoTransformer & 0.79\\
HPG-Diff & 2.79\\
TopoMamba & 0.48\\
\hline
\end{tabular}
\end{table}

\subsection{Ablation Studies}\label{sec:ablation}

We conduct controlled ablation studies to examine the effectiveness of the key architectural designs in TopoMamba. In addition to removing the physical-field input as a reference, we focus on the components directly associated with our main methodological contributions, including physical fields, the explicit load-support relation representation, and adaptive feature fusion. The explicit relation representation is designed to provide the model with structured mechanical dependencies between loads and supports, while the multi-directional scanning mechanism facilitates long-range dependency modeling along different spatial directions. The adaptive fusion module further enables these directional features to be dynamically integrated according to the underlying mechanical conditions. As shown in Table~\ref{tab:ablation}, removing any of these components consistently degrades performance, with a more evident deterioration under the OOD tests. This trend indicates that the proposed designs provide benefits: explicit relation modeling introduces mechanically meaningful structural priors, multi-directional scanning improves long-range spatial dependency modeling, and adaptive fusion enhances the model's ability to adjust feature aggregation across varying load and support configurations. Together, these components contribute to the robustness and generalization capability of TopoMamba under distribution shifts.

\begin{table}[!htb]
\caption{Ablation study on ID and OOD tests.}
\label{tab:ablation}\centering
\begin{tabular}{ccccc}
\hline
\shortstack{Mean CE\\(\%)} &
\shortstack{TopoMamba\\(5SIMP)} &
\shortstack{w/o\\Physical Fields} &
\shortstack{w/o Load--support\\Relations} &
\shortstack{w/o Adaptive\\Fusion} \\
\hline
ID   & \textbf{0.43} & 1.24  & 0.49 & 0.47 \\
OOD  & \textbf{1.79} & 10.00 & 1.83 & 1.91 \\
\hline
\end{tabular}
\end{table}
\section{Conclusion}
This paper presents TopoMamba, a load-support relation-guided multi-directional state-space framework for topology prediction. TopoMamba achieves strong predictive performance and generalization with a compact parameterization and low computational cost, outperforming existing state-of-the-art methods on the benchmarks. These results demonstrate the potential of deterministic learning-based models to provide an effective balance between structural quality, generalization, and inference efficiency, and offer a promising direction for accelerating TO in high-dimensional design spaces and computationally intensive iterative settings.

Although our methods perform well on current benchmarks, their generalization to more complex design conditions requires further improvement. Future work should enhance generalization while preserving low inference cost, and extend the framework beyond 2D regular grids toward irregular discretizations, 3D structures.
\bibliographystyle{unsrt}  
\bibliography{references}

\end{document}